\documentclass[letterpaper, 10 pt, conference]{ieeeconf}  
\IEEEoverridecommandlockouts                            
\usepackage{url}
\usepackage[utf8]{inputenc}
\usepackage[T1]{fontenc}
\usepackage{amsmath,amssymb,amsfonts}
\usepackage{graphicx}
\usepackage{booktabs}
\usepackage{array}
\usepackage{algorithm}
\usepackage{algorithmic}
\usepackage{xcolor}
\usepackage{multirow}

\usepackage{xcolor}

\title{\LARGE \bf Can a Robot Read Braille? — Learning to Adapt Contact via Imitation Learning for Tactile Braille Recognition}

\author{
Xi Chen$^{1,2}$, Yunlong Shan$^{3}$, Sihan Chen$^{4}$, Jun Hu$^{2,5}$, Zhongxuan Li$^{4}$,\\
Shiyao Zhang$^{2}$, Sichao Liu$^{6}$, Zhong Zhao$^{1}$, Kosta Jovanovi\'c$^{7}$, and Peng Zhou$^{2,*}$\\[1mm]
{\small $^{1}$College of Mechatronics and Control Engineering, Shenzhen University, Shenzhen, Guangdong, China}\\
{\small $^{2}$School of Advanced Engineering, Great Bay University, Dongguan, Guangdong, China}\\
{\small $^{3}$Jiangsu Key Laboratory of Drug Metabolism and Pharmacokinetics, China Pharmaceutical University, Nanjing, Jiangsu, China}\\
{\small $^{4}$Department of Computer Science, School of Computing and Data Science, The University of Hong Kong, Pokfulam, Hong Kong SAR, China}\\
{\small $^{5}$Tsinghua University, Beijing, China}\\
{\small $^{6}$\'Ecole Polytechnique F\'ed\'erale de Lausanne (EPFL), Lausanne, Switzerland}\\
{\small $^{7}$School of Electrical Engineering, University of Belgrade, Belgrade, Serbia}\\
{\small $^{*}$Corresponding author.}
}

\begin{document}

\maketitle
\thispagestyle{empty}
\pagestyle{empty}

\begin{abstract}
For people who are blind, touch provides an essential channel for accessing written information through Braille. Bringing a similar capability to robots requires them not only to recognize tactile patterns, but also to actively establish physical contact that makes those patterns readable. Yet existing robotic Braille readers largely focus on recognition after contact, leaving contact establishment itself insufficiently addressed. We present an adaptive-contact framework for robotic tactile Braille reading that assesses contact quality and physically corrects unsuitable contact before recognition and reconstruction. Multi-Head Policy Learning uses expert-guided contact-adjustment demonstrations to jointly learn contact acceptability and pose corrections. During deployment, the robot iteratively evaluates and re-establishes contact, retaining reliable tactile observations for pose-aware fusion and Braille reconstruction. Across 20 physical Braille plates used for learning and evaluation, the proposed approach achieves 94.0\% tactile quality and 88.6\% tactile reconstruction on the ten online-evaluation plates. These results demonstrate the importance of actively establishing readable contact, rather than relying solely on recognition under imperfect tactile observations, for reliable robotic Braille reading. 

\end{abstract}


\section{Introduction}
\begin{figure}[t]
    \centering
    \includegraphics[width=0.98\columnwidth,
    trim=7.2cm 5.5cm 8.2cm 1.5cm,
    clip
    ]{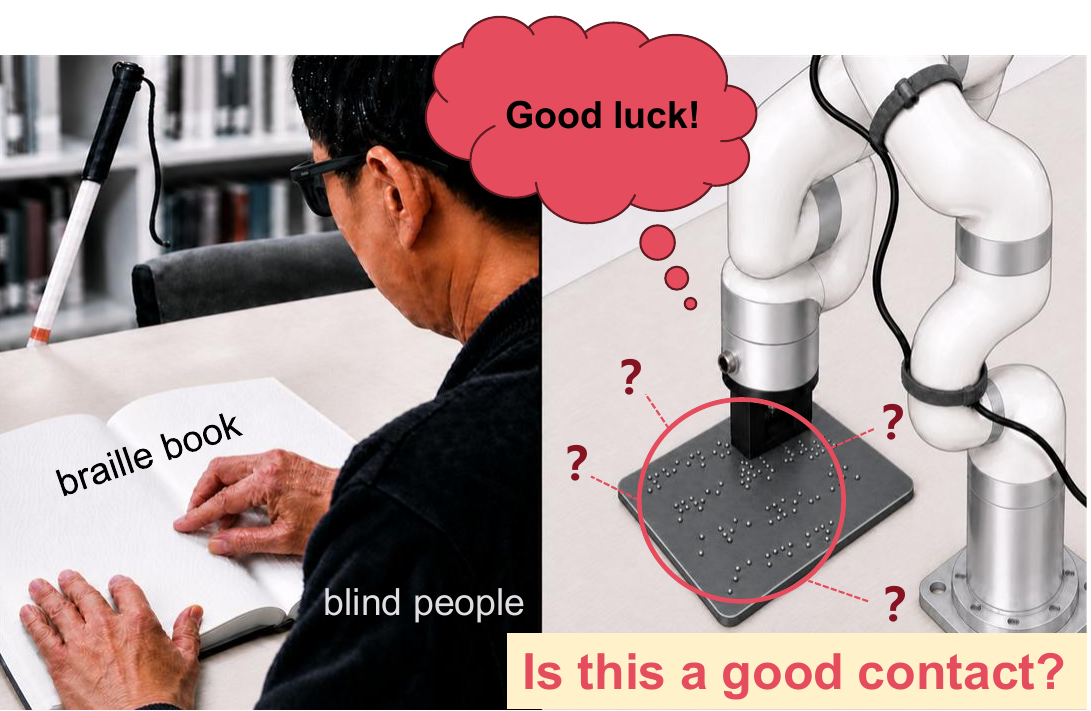}
    \caption{Motivation for adaptive contact in robotic tactile Braille reading: humans naturally adjust fingertip contact, while robots must actively establish readable tactile contact.}
    \label{fig:system}
\end{figure}

Braille provides blind people with a tactile means of accessing written information embedded in everyday environments. Enabling assistive and service robots to access the same tactile information would allow them to interact with Braille-encoded objects and interfaces designed for human use (Fig.~\ref{fig:system}). Beyond this practical motivation, Braille presents a representative challenge for robotic active touch: its information is encoded by small, structured surface features whose tactile appearance depends strongly on how physical contact is established. Unlike visual text recognition, robotic Braille reading therefore requires not only interpreting sensory observations, but actively interacting with the surface to obtain readable tactile evidence. Insufficient or excessive indentation can suppress or merge dots, while sensor tilt can spatially distort their geometry. Reliable robotic Braille reading consequently begins with a fundamental interaction problem: how can a robot establish physical contact that makes Braille readable?

Robotic tactile Braille reading has been explored with skin-like~\cite{zhao2020skin}, magnetic~\cite{alfadhel2016magnetic}, and neuromorphic tactile sensors~\cite{xu2025neuromorphic} under pressing, tapping, and sliding acquisition~\cite{potdar2024highspeed}. Recent systems have improved reading speed through sliding interactions~\cite{potdar2024highspeed} and evaluated Braille recognition across variations in tapping speed, contact position, and indentation depth~\cite{xu2025neuromorphic}. Active approaches have also used recognition outputs to adapt scanning motion~\cite{bologna2012active,bologna2013closedloop}. However, these methods primarily focus on recognizing tactile observations after acquisition. Contact errors can directly corrupt the geometric evidence before recognition: insufficient indentation may suppress dots, excessive indentation may merge them, and sensor tilt may cause spatially uneven deformation (Fig.~\ref{fig:system}). Rather than requiring a recognizer to tolerate such degraded observations, we ask a complementary question: \emph{can a robot actively establish readable contact before
recognition?}

To address this problem, we propose an image-conditioned adaptive-contact framework for robotic tactile Braille reading. Multi-Head Policy Learning uses expert-guided contact-adjustment demonstrations to jointly learn contact acceptability and pose corrections. During deployment, the robot iteratively evaluates and corrects unsuitable contact, retaining selected tactile observations together with their measured acquisition poses. These observations are then registered through pose-aware overlap fusion and dot refinement to reconstruct global Braille. In this way, the proposed framework connects contact assessment, physical correction, and tactile reconstruction in a closed-loop Braille-reading pipeline.

The main contributions of this work are threefold:
\begin{itemize}
    \item We formulate \textbf{adaptive contact establishment for
    robotic tactile Braille reading}, in which the robot explicitly
    assesses whether physical contact preserves readable Braille-dot
    geometry and re-establishes unsuitable contact before
    reconstruction.

    \item We develop a \textbf{Multi-Head Policy Learning approach}
    from grouped expert-guided demonstrations. The approach jointly
    predicts contact acceptability and pose corrections,
    enabling closed-loop recovery from contact errors.

    \item We integrate adaptive contact with \textbf{pose-aware
    multi-view tactile reconstruction} and validate the complete
    pipeline on physical Chinese and English Braille plates. The
    proposed full correction achieves 94.0\% tactile quality and
    88.6\% tactile reconstruction.
\end{itemize}

\section{Related Work}
\label{sec:related_work}

\subsection{Robotic Tactile Braille Reading and Active Contact}

Active perception couples purposeful motion~\cite{bajcsy1988active,prescott2011active,seminara2019active} with sensory evidence~\cite{li2020review}, and human haptic exploration and robotic active touch similarly link action to recognition~\cite{lederman1987hand,lepora2013activebayesian}. Vision-based tactile sensors provide image-based observations of contact geometry and deformation~\cite{yuan2017gelsight,zhou2026vision}.

Robotic Braille reading has been studied using neurorobotic~\cite{bologna2011encoding}, skin-like~\cite{zhao2020skin}, visuo-tactile~\cite{zhang2025braille}, magnetic~\cite{alfadhel2016magnetic}, neuromorphic~\cite{xu2025neuromorphic,mullercleve2022braille}, and dual-mode sensing~\cite{gao2024enhanced} under static, tapping, and sliding interactions. Recent systems have extended reading speed~\cite{potdar2024highspeed}, compactness~\cite{jenkinson2024brailletip}, and operating conditions~\cite{xu2025neuromorphic,zhang2025braille}. Active robotic readers have also regulated scanning velocity and trajectory using recognition feedback~\cite{bologna2012active,bologna2013closedloop}. However, these methods primarily prescribe contact through the acquisition protocol or treat contact variation as a recognition condition, rather than explicitly assessing Braille-dot readability and correcting contact position and orientation before reconstruction.

\subsection{Tactile Servoing and Reconstruction}

Tactile control demonstrates that physical contact can be actively regulated. Prior work has controlled contact position, force, and edge orientation~\cite{li2013control}, learned image-space dynamics for model-predictive manipulation~\cite{tian2019manipulation}, and used pose and shear feedback to regulate contact depth and orientation~\cite{lepora2021pose,lloyd2024pose}. Tactile histories have supported imitation under partial observability~\cite{yang2023seq2seq}. These methods establish useful contact-control principles, but do not target Braille-dot readability.

Tactile mapping provides a complementary basis for reconstruction. Existing approaches combine tactile observations with robot kinematics~\cite{bauza2019mapping}, jointly estimate shape and pose~\cite{suresh2021tactileslam}, incorporate vision and loop closure~\cite{zhao2023fingerslam}, fuse local tactile geometry with vision~\cite{suresh2022shapemap,xu2023vtaco}, or disentangle shear for planar reconstruction~\cite{gupta2022disentanglement}. Their focus is primarily on registration and fusion after contact acquisition. In contrast, our method links Braille-specific contact-quality assessment and pose correction with tactile reconstruction, allowing us to re-establish unsuitable contact before retaining an observation.

\section{Problem Formulation}
\label{sec:problem}

\begin{figure}[t]
    \centering
    \includegraphics[width=\columnwidth,
    trim=4.0cm 3.8cm 10cm 3.2cm,
    clip
    ]{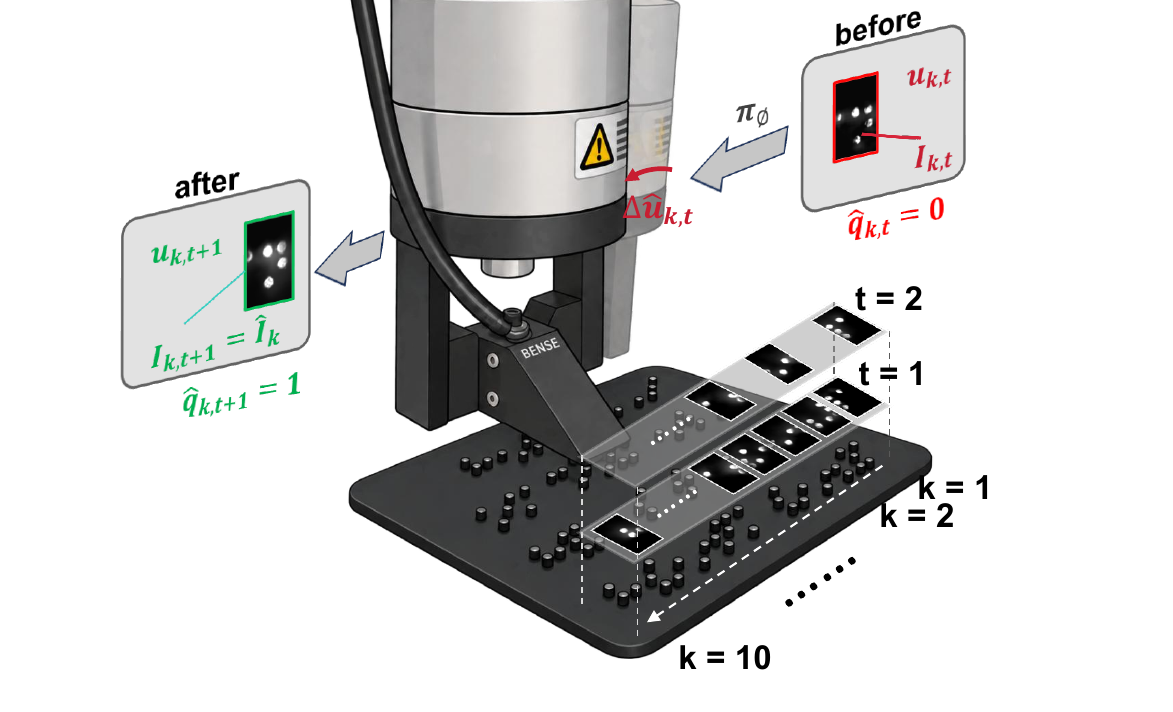}
    \caption{Problem formulation for adaptive contact. At trajectory location \(k\), the policy \(\pi_{\phi}\) evaluates the tactile image \(I_{k,t}\) acquired at contact pose \(\mathbf u_{k,t}\). An unsuitable contact is rejected and the predicted correction \(\Delta\hat{\mathbf u}_{k,t}\) updates the next pose; a suitable tactile image is retained as \(\hat I_k\).}
    \label{fig:problem_formulation}
\end{figure}

The task is to read a sequence of Braille characters by moving an end-effector-mounted tactile sensor along the Braille surface and acquiring tactile images through contact. As illustrated in Fig.~\ref{fig:problem_formulation}, the robot follows a predefined trajectory of \(k\) ordered lateral locations separated by a fixed step, covering the Braille line from beginning to end. At contact attempt \(t\) at location \(k\), the commanded end-effector pose is \(\mathbf u_{k,t}=[\mathbf p_{k,t}^{\mathsf T},\boldsymbol{\theta}_{k,t}^{\mathsf T}]^{\mathsf T}\), where \(\mathbf p_{k,t}\) and \(\boldsymbol{\theta}_{k,t}\) denote position and orientation, respectively. The resulting tactile image is denoted by \(I_{k,t}\), where \(t\) indexes repeated contact attempts when the observed dot geometry is unsuitable, e.g., missing, incomplete, merged, or deformed.

We formulate local contact adaptation as a partially observable Markov decision process (POMDP) \(\mathcal{M}=(\mathcal{S},\mathcal{A},\mathcal{O},\mathcal{T},\mathcal{Z},\mathcal{R})\). The latent state \(s_{k,t}\in\mathcal{S}\) represents the underlying sensor--surface contact configuration, including the relative contact pose and local deformation that determine Braille-dot geometry. This state is not directly observable; instead, the robot receives a tactile observation \(o_{k,t}=I_{k,t}\in\mathcal{O}\). The action \(a_{k,t}\in\mathcal{A}\) consists of either accepting the current contact or applying a pose correction \(\Delta\mathbf u_{k,t}\) before the next attempt. The transition model \(\mathcal{T}\) describes how the contact state changes after pose adjustment, while the observation model \(\mathcal{Z}\) relates the latent contact state to the acquired tactile image. The objective is to obtain readable tactile observations while minimizing unnecessary re-contact attempts.
Rather than explicitly estimating a belief state and solving the full POMDP, we learn an observation-conditioned reactive policy from expert-guided demonstrations:
\begin{equation}
(\hat q_{k,t},\Delta\hat{\mathbf u}_{k,t})
=\pi_{\phi}(I_{k,t}),
\label{eq:policy}
\end{equation}
where \(\pi_{\phi}\) is parameterized by \(\phi\). The contact-quality score \(\hat q_{k,t}\in[0,1]\) estimates whether the current tactile observation preserves usable Braille-dot geometry, while \(\Delta\hat{\mathbf u}_{k,t}\) predicts the pose correction for the next contact attempt. The score is compared with an acceptance threshold \(\tau_q\). If \(\hat q_{k,t}\geq\tau_q\), \(I_{k,t}\) is retained as \(\hat I_k\); otherwise, the robot updates its contact pose and performs another attempt.

\begin{figure*}[t]
    \centering
    \includegraphics[width=0.99\textwidth,
    trim=1.0cm 2.5cm 0.6cm 2.0cm,
    clip
    ]{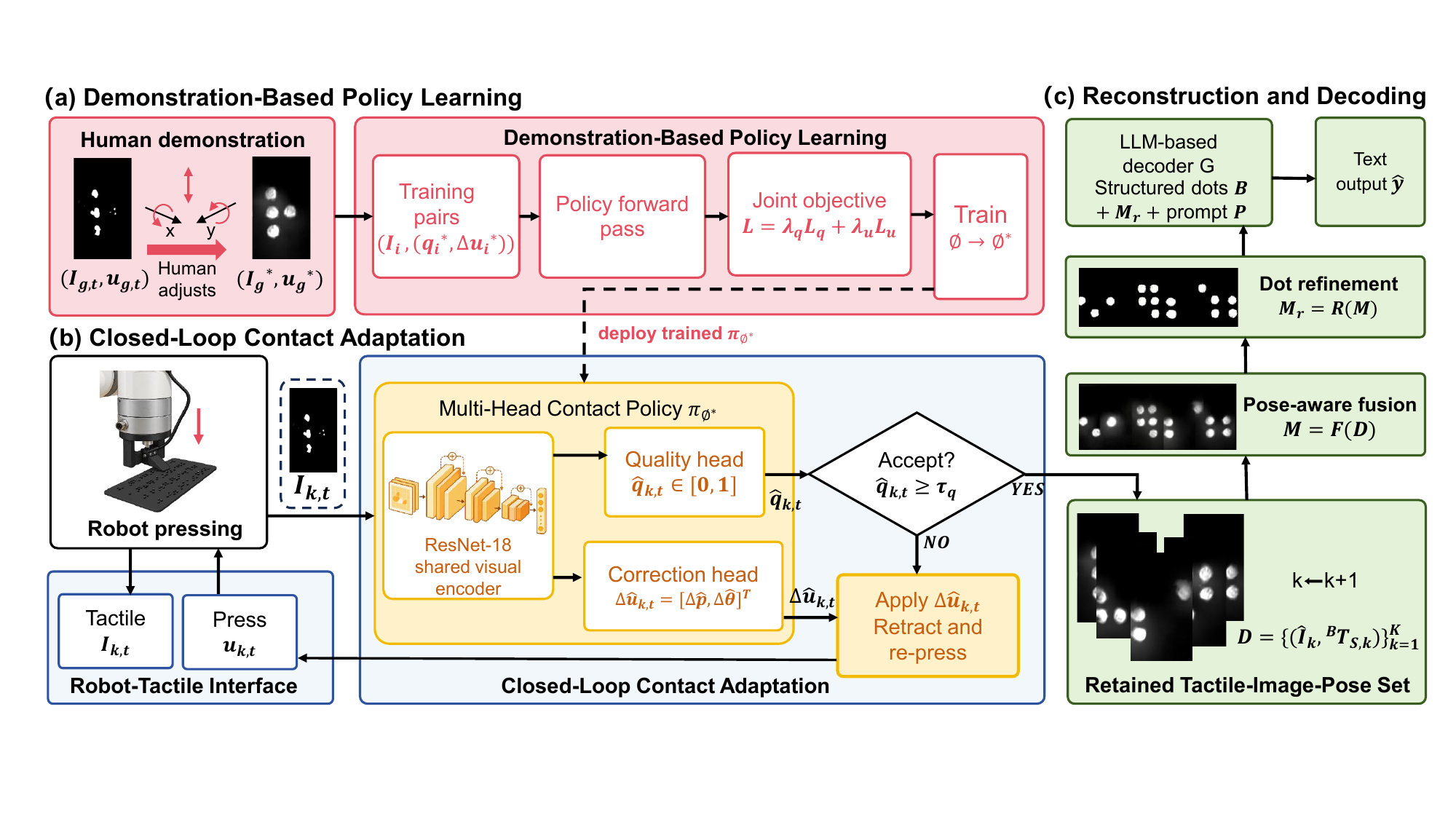}
\caption{Overview of the proposed method. (a) Expert contact-adjustment demonstrations are collected and converted into supervision for Multi-Head Policy Learning. (b) The learned policy is deployed in closed loop to improve unsuitable contacts. (c) Reconstruction and decoding fuse retained tactile-image--pose pairs, refine the resulting Braille dots, and decode the text.}
    \label{fig:framework}
\end{figure*}

\section{Methodology}
\label{sec:method}
\subsection{Method Overview}
Figure~\ref{fig:framework} provides an overview of the proposed method, comprising three stages. First, expert contact-adjustment demonstrations are collected and converted into supervision for Multi-Head Policy Learning [Fig.~\ref{fig:framework}(a)]. Second, the resulting learned policy is deployed in a closed loop to assess each tactile image $I_{k,t}$ and either retain it as $\hat I_k$ or correct the contact pose $\mathbf u_{k,t}$ and re-contact the surface [Fig.~\ref{fig:framework}(b)]. Finally, the retained tactile-image--pose pairs are fused into a global tactile image $M$, refined into $M_r$, and organized into Braille cells for text decoding [Fig.~\ref{fig:framework}(c)]. A Braille cell consists of six dot positions arranged in two columns of three; hereafter, \emph{dot} denotes an individual position and \emph{cell} denotes the complete six-position unit.

\subsection{Learning Adaptive Contact from Expert Demonstrations}
\subsubsection{Expert Demonstration and Target Construction}
\begin{figure}[t]
    \centering
    \includegraphics[width=\columnwidth,
    trim=5cm 5.0cm 5cm 3cm,
clip
    ]{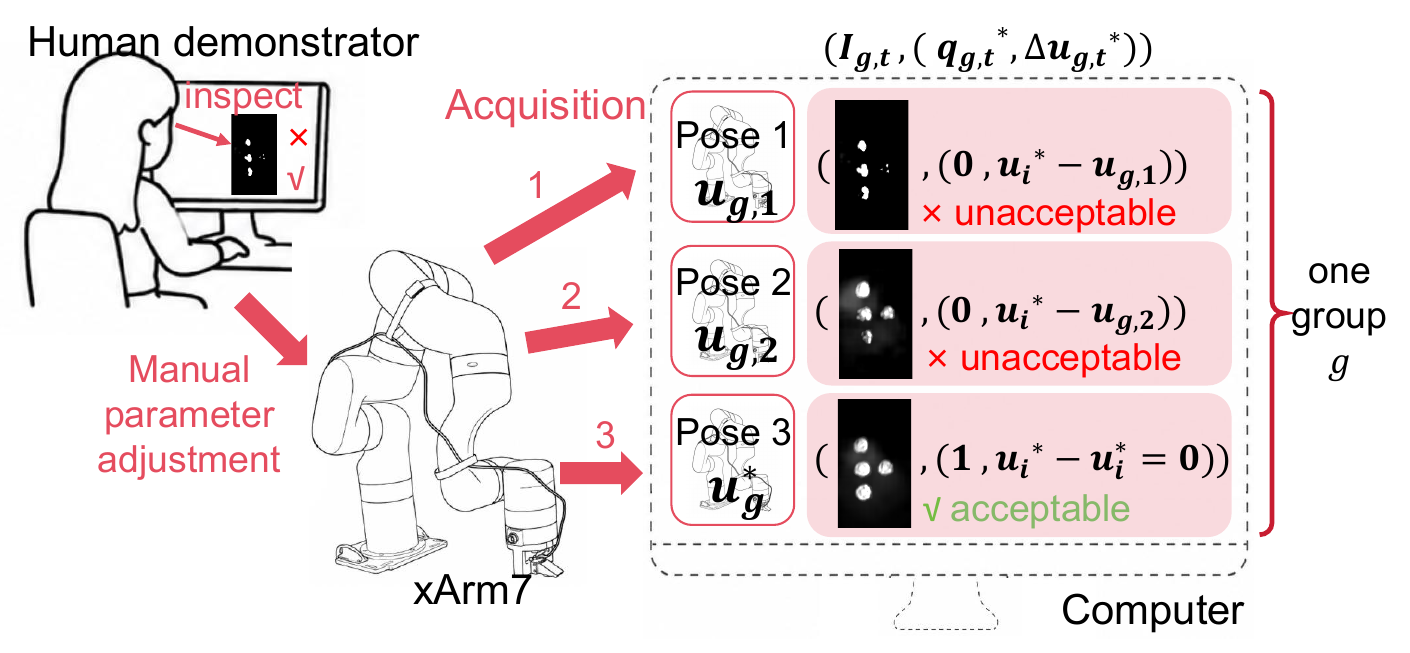}
    \caption{Construction of one grouped expert demonstration. The expert inspects each tactile image $I_{g,t}$ and changes $\mathbf p_{g,t}$ and $\boldsymbol\theta_{g,t}$ through UFACTORY Studio until the contact is accepted. Each intermediate tactile image becomes a rejected sample whose pose-correction target points directly to the final accepted end-effector pose $\mathbf u_g^{*}$; the final tactile image $I_g^{*}$ becomes an accepted sample with a zero correction target.}
    \label{fig:demonstration_collection}
\end{figure}

As illustrated in Fig.~\ref{fig:demonstration_collection}, expert demonstrations are collected offline at selected Braille locations. At each location, a demonstration group \(g\) consists of a sequence of contact attempts indexed by \(t\). Starting from a nominal pose, the robot acquires a tactile image \(I_{g,t}\) at each attempt, and the expert assesses whether the Braille-dot geometry is readable. Contacts exhibiting missing, incomplete, merged, or deformed dots are rejected, and the expert adjusts the contact pose using a teach pendant until an acceptable contact is obtained. The final accepted image and contact pose are denoted by \((I_g^{*},\mathbf u_g^{*})\).

Each attempt is assigned a binary contact-quality target \(q_{g,t}^{*}\), where \(q_{g,t}^{*}=1\) for the final accepted contact and \(q_{g,t}^{*}=0\) otherwise. For a rejected attempt, the correction target is defined as the residual from its current contact parameters to the final accepted ones:
\begin{equation}
\Delta\mathbf u_{g,t}^{*}=\mathbf u_g^{*}-\mathbf u_{g,t},
\label{eq:demonstration_target}
\end{equation}
where \(\mathbf u=[\mathbf p^{\mathsf T},\boldsymbol\theta^{\mathsf T}]^{\mathsf T}\) contains the contact position \(\mathbf p\) and orientation \(\boldsymbol\theta\). For the accepted contact, \(\Delta\mathbf u_{g,t}^{*}=\mathbf 0\). This grouped construction allows multiple rejected contacts to share the same accepted target; for example, two rejected attempts followed by an accepted one yield two distinct correction samples toward the accepted contact. Each recorded attempt therefore forms a supervised training pair \(\left(I_{g,t},(q_{g,t}^{*},\Delta\mathbf u_{g,t}^{*})\right)\).

\subsubsection{Contact-Quality and Pose-Correction Policy}

For policy learning, the input--target pairs from all demonstration groups are flattened into training samples indexed by \(i\). Given a tactile image \(I_i\), the policy \(\pi_{\phi}\) uses a shared ImageNet-pretrained ResNet-18 encoder followed by two heads: a contact-quality head predicting \(\hat q_i\in[0,1]\) and a pose-correction head predicting \(\Delta\hat{\mathbf u}_i=[(\Delta\hat{\mathbf p}_i)^{\mathsf T},(\Delta\hat{\boldsymbol\theta}_i)^{\mathsf T}]^{\mathsf T}\). To accommodate the single-channel tactile input, the first-layer RGB weights are averaged across channels. Each tactile image is converted to a \(700\times400\) single-channel 8-bit image using 2nd--98th percentile contrast mapping. The correction components are normalized by their standard deviations over rejected training samples, collected in \(\mathbf s\) and fixed after training-data preprocessing.

The quality and correction heads are trained jointly. For a mini-batch of \(N\) tactile images, the objective is
\begin{equation}
\begin{aligned}
\mathcal L(\phi)&=\lambda_q\mathcal L_q+\lambda_u\mathcal L_u,\\
\mathcal L_q&=\frac{1}{N}\sum_{i=1}^{N}\operatorname{BCE}(\hat q_i,q_i^{*}),\\
\mathcal L_u&=\frac{1}{N_-}\sum_{i=1}^{N}(1-q_i^{*})\,
\mathrm{SL1}_{\beta}\!\left(
\frac{\Delta\hat{\mathbf u}_i-\Delta\mathbf u_i^{*}}{\mathbf s}\right),
\end{aligned}
\label{eq:loss}
\end{equation}
where \(N_-=\max\!\left(1,\sum_{i=1}^{N}(1-q_i^{*})\right)\), and \(\lambda_q\) and \(\lambda_u\) balance contact-quality classification and pose-correction regression. The correction loss is computed only for rejected contacts (\(q_i^{*}=0\)), since accepted contacts require no further correction. Mini-batches are sampled by demonstration group with an approximately \(1{:}1\) accepted-to-rejected ratio, while the loss weights and model selection are determined using validation data.

\begin{algorithm}[t]
\caption{Closed-Loop Contact Adaptation}
\label{alg:closed_loop}
\small
\begin{algorithmic}[1]
\REQUIRE Trajectory locations $1{:}K$, policy $\pi_{\phi}$, $\tau_q=0.80$, $T_{\max}=3$
\ENSURE Retained tactile-image--pose set $\mathcal D$
\STATE $\mathcal D\leftarrow\emptyset$
\STATE Initialize contact pose $\mathbf u$ from the nominal pose
\FOR{$k=1$ to $K$}
    \STATE Initialize $\mathbf u_{k,1}$ using the inherited contact parameters
    \STATE $\mathrm{accepted}\leftarrow\mathrm{false}$; initialize best-scoring attempt
    \FOR{$t=1$ to $T_{\max}$}
        \STATE Establish contact at $\mathbf u_{k,t}$ and wait for stabilization
        \STATE Acquire $I_{k,t}$ and ${}^{B}\mathbf T_{S,k,t}$
        \STATE $(\hat q_{k,t},\Delta\hat{\mathbf u}_{k,t})\leftarrow\pi_{\phi}(I_{k,t})$
        \STATE Update the best-scoring attempt using $\hat q_{k,t}$
        \STATE Retract the sensor
        \IF{$\hat q_{k,t}\geq\tau_q$}
            \STATE Retain $(\hat I_k,{}^{B}\mathbf T_{S,k})\leftarrow(I_{k,t},{}^{B}\mathbf T_{S,k,t})$
            \STATE Inherit the accepted contact parameters for location $k+1$
            \STATE $\mathrm{accepted}\leftarrow\mathrm{true}$; \textbf{break}
        \ELSIF{$t<T_{\max}$}
            \STATE $\mathbf u_{k,t+1}\leftarrow\mathbf u_{k,t}+\Delta\hat{\mathbf u}_{k,t}$
            \STATE Bound the contact pose to the tested safety range
        \ENDIF
    \ENDFOR
    \IF{$\mathrm{accepted}=\mathrm{false}$}
        \STATE Retain the tactile-image--pose pair with the highest $\hat q_{k,t}$
    \ENDIF
    \STATE $\mathcal D\leftarrow\mathcal D\cup\{(\hat I_k,{}^{B}\mathbf T_{S,k})\}$
\ENDFOR
\end{algorithmic}
\end{algorithm}

\subsection{Closed-Loop Contact Adaptation}

During deployment, the learned policy is executed in a closed loop at each trajectory location \(k\). Starting from the nominal contact pose, the robot establishes contact, waits for stabilization, and acquires a tactile image \(I_{k,t}\) together with the corresponding sensor pose \({}^{B}\mathbf T_{S,k,t}\). The policy then predicts the contact-quality score \(\hat q_{k,t}\) and pose correction \(\Delta\hat{\mathbf u}_{k,t}\). If \(\hat q_{k,t}\geq\tau_q\), the contact is accepted and \(I_{k,t}\) is retained as \(\hat I_k\). Otherwise, the robot retracts and updates the contact pose according to
\begin{equation}
\mathbf u_{k,t+1}=\mathbf u_{k,t}+\Delta\hat{\mathbf u}_{k,t},
\label{eq:online_update}
\end{equation}
followed by another contact attempt. The updated contact pose is bounded to the tested safety range. The acceptance threshold and maximum number of attempts are fixed to \(\tau_q=0.80\) and \(T_{\max}=3\), respectively.

Each attempt is associated with \((I_{k,t},{}^{B}\mathbf T_{S,k,t},\hat q_{k,t})\), where \({}^{B}\mathbf T_{S,k,t}\) is the measured transformation from the sensor frame \(S\) to the robot-base frame \(B\). If no attempt satisfies the acceptance criterion within \(T_{\max}\), the tactile-image--pose pair with the highest \(\hat q_{k,t}\) is retained and location \(k\) is marked unsuccessful. For the retained observation, we omit the attempt index and denote the pair by \((\hat I_k,{}^{B}\mathbf T_{S,k})\). The retained pairs are subsequently used for pose-aware reconstruction in Sec.~\ref{subsec:pose_fusion}. The complete procedure is summarized in Alg.~\ref{alg:closed_loop}.

\subsection{Pose-Aware Braille Reconstruction}
\label{subsec:pose_fusion}

After closed-loop contact adaptation, one tactile image and its measured sensor pose are retained at each trajectory location, forming
\begin{equation}
\mathcal D=
\left\{\left(\hat I_k,{}^{B}\mathbf T_{S,k}\right)\right\}_{k=1}^{K},
\label{eq:retained_set}
\end{equation}
where ${}^{B}\mathbf T_{S,k}$ denotes the transformation from the sensor frame \(S\) to the robot-base frame \(B\). Since each \(\hat I_k\) captures only a local region of the Braille plate, the retained observations are registered in a common plate coordinate frame using their measured sensor poses. Specifically, each tactile image is binarized using a validation-selected threshold of 156, and the four corners of its \(24.5\times15.0\)~mm sensing region are projected onto the Braille-plate plane according to ${}^{B}\mathbf T_{S,k}$. The resulting pose-aware fusion is denoted by
\begin{equation}
M=\mathcal F(\mathcal D),
\label{eq:reconstruction}
\end{equation}
where \(M\) is the reconstructed global tactile image. To reduce unreliable boundary responses, each valid tile region is eroded by two pixels, Gaussian-smoothed with a one-pixel standard deviation, and thresholded at 0.02; overlapping valid regions are fused by retaining the maximum warped response.

\subsubsection{Dot Refinement and Braille Decoding}
The fused image \(M\) may contain spurious, incomplete, or merged dot responses, particularly near tile boundaries. We therefore apply a refinement operator \(R(\cdot)\) that removes spurious components, regularizes incomplete dots to the nominal circular shape, and separates merged dots:
\begin{equation}
M_r=R(M),
\label{eq:refinement}
\end{equation}
where \(M_r\) denotes the refined Braille-dot image. All refinement thresholds are selected using validation data and fixed before testing.

The detected dot centers in \(M_r\) are organized into ordered Braille lines and cells using OpenCV. Each cell contains six possible dot positions, with positions 1--3 ordered from top to bottom in the left column and positions 4--6 in the right column. Their binary occupancy states form the structured Braille-cell sequence \(\mathcal B\).

For semantic decoding, a language-specific decoder \(G\) receives the refined dot image \(M_r\), the structured cell sequence \(\mathcal B\), and a fixed prompt \(P\):
\begin{equation}
\hat{\mathbf y}=G(M_r,\mathcal B;P),
\label{eq:recognition}
\end{equation}
where \(\hat{\mathbf y}\) denotes the decoded text. We implement \(G\) using a large language model (LLM), with separate fixed prompts for Chinese and English and identical decoder settings across all acquisition strategies. Each Braille line is decoded once. The decoder is used only for downstream text interpretation; tactile quality and tactile reconstruction are computed independently of \(G\).

\begin{figure}[t]
\centering
\includegraphics[width=0.98\columnwidth,
trim=6.7cm 5.4cm 11cm 3cm,
clip
]{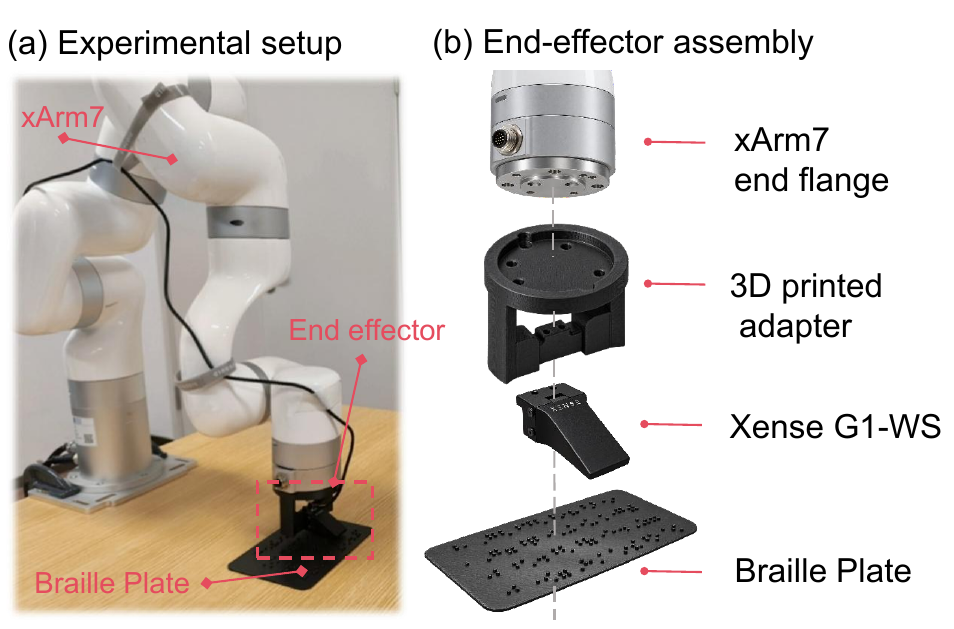}
\caption{Experimental platform and end-effector assembly. (a) The xArm7 establishes tactile contact with a Braille plate. (b) Exploded view of the xArm7 end flange, custom 3-D-printed adapter, Xense G1-WS tactile sensor, and Braille plate.}
\label{fig:experimental_setup}
\end{figure}



\begin{table}[t]
\caption{Offline evaluation on held-out Braille plates.}
\label{tab:offline_policy_evaluation}
\centering
\footnotesize
\setlength{\tabcolsep}{10pt}
\renewcommand{\arraystretch}{1.02}
\begin{tabular}{@{}lcc@{}}
\toprule
\textbf{Task} & \textbf{Metric} & \textbf{Value} \\
\midrule
\multirow{2}{*}{Quality}
    & F1 (\%) & 92.8 \\
    & Error (\%) & 5.9 \\
\addlinespace[1pt]
\multirow{4}{*}{Correction}
    & Pos. MAE (mm) & 0.20 \\
    & Roll MAE ($^\circ$) & 0.27 \\
    & Pitch MAE ($^\circ$) & 0.29 \\
    & Mean $R^2$ & 0.76 \\
\bottomrule
\end{tabular}
\end{table}

\begin{figure}[t]
\centering
\includegraphics[width=0.98\columnwidth,
trim=4.5cm 4.0cm 6.0cm 3cm,
clip
]{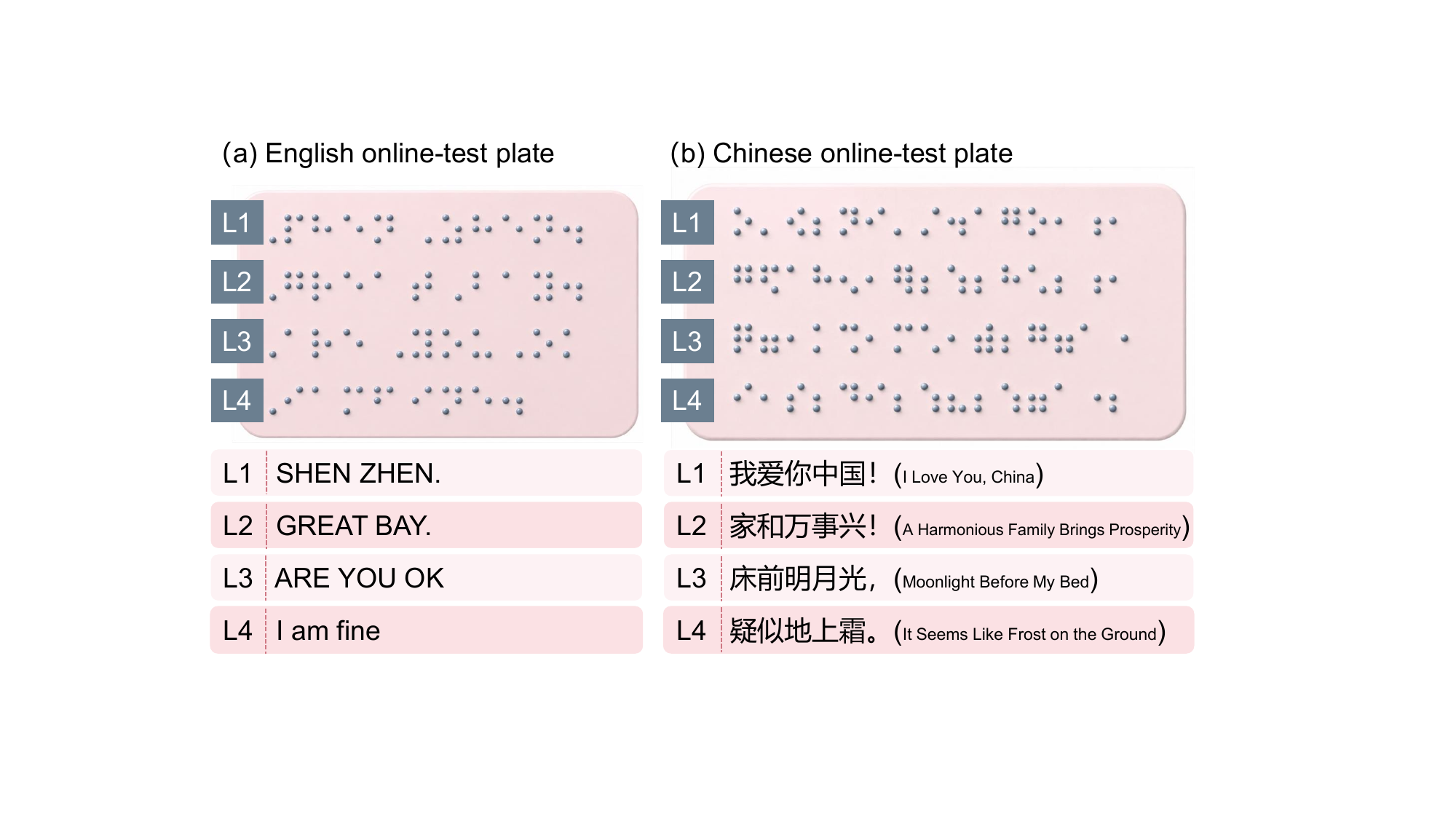}
\caption{Representative Braille plates reserved for online evaluation and their ground-truth line transcriptions. (a) English Braille plate with 10 trajectory locations per line. (b) Chinese Braille plate with 13 trajectory locations per line. L1--L4 run from top to bottom; parenthetical English translations are annotations, not decoder outputs.}
\label{fig:online_test_plates}
\end{figure}

\begin{figure*}[t]
    \centering
    \includegraphics[width=0.99\textwidth]{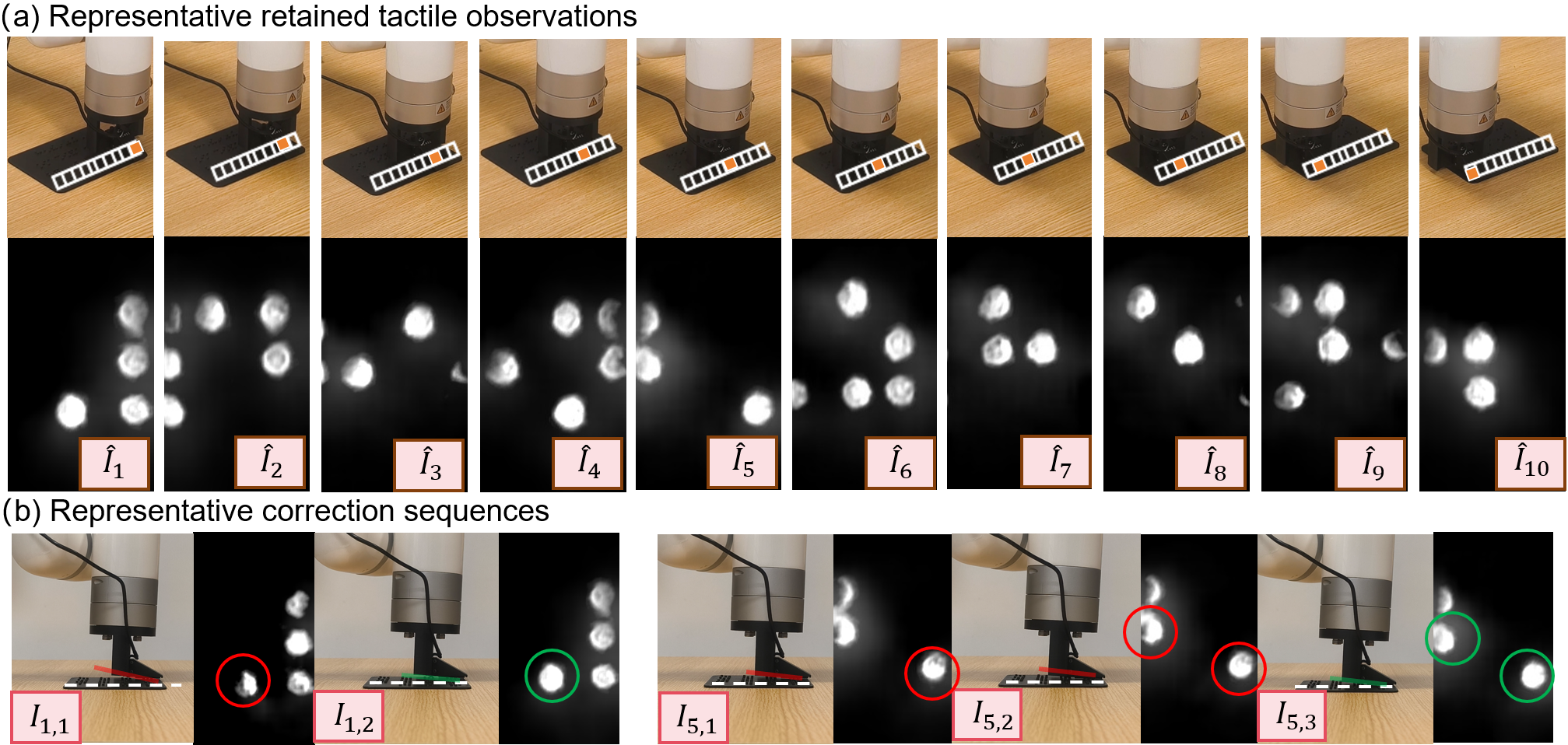}
\caption{Representative acquisition with the Full strategy. (a) Retained tactile images $\hat I_1$--$\hat I_{10}$ at ten consecutive trajectory locations; orange markers indicate the current trajectory location on the Braille plate. (b) Contact-attempt sequences at trajectory locations $k=1$ and $k=5$. Tactile images $I_{1,1}$, $I_{5,1}$, and $I_{5,2}$ are rejected (red), whereas $I_{1,2}$ and $I_{5,3}$ are accepted after pose correction (green).}
    \label{fig:adaptive_acquisition}
\end{figure*}

\section{Experiments and Results}
\label{sec:experiments}

The experiments evaluate two questions: (1) whether the learned policy can assess contact quality and predict corrective motion on unseen Braille plates, and (2) whether closed-loop contact adaptation improves tactile acquisition and downstream Braille reconstruction and reading.

\subsection{Experimental Setup}
\label{subsec:exp_setup}

\subsubsection{Platform and Dataset}

The experimental platform consists of an xArm7 robot equipped with an Xense G1-WS optical tactile sensor through a custom 3-D-printed adapter, as shown in Fig.~\ref{fig:experimental_setup}. We use 20 black Braille plates, equally divided between Chinese and English. Ten plates (five Chinese and five English) are used for offline demonstration collection, while the remaining ten are reserved for online evaluation.

For offline data collection, the expert selects 10 contact locations on each of four lines per plate, yielding 40 demonstration groups per collection. The dataset is split by Braille plate to prevent tactile patterns from the same plate appearing across training and evaluation: four plates collected three times provide 480 training groups, two plates collected three times provide 240 validation groups, and four plates collected twice provide 320 held-out test groups. In total, the 1,040 groups contain 2,426 tactile images, comprising 1,040 accepted and 1,386 rejected contacts.

\subsubsection{Policy Training and Offline Evaluation}

The Multi-Head policy is trained using AdamW for 50 epochs with a learning rate and weight decay of \(1\times10^{-4}\), a mini-batch size of 16, \(\lambda_q=\lambda_u=1\), and \(\beta=1.0\). The checkpoint with the lowest validation joint loss is selected, and all reported offline results are computed on the held-out test plates. A contact is predicted as acceptable when \(\hat q_i\geq\tau_q=0.80\).

We evaluate contact-quality assessment using F1 score and classification error, and pose correction using mean absolute error (MAE) and mean \(R^2\) over rejected contacts. As shown in Table~\ref{tab:offline_policy_evaluation}, the policy achieves a contact-quality F1 of 92.8\% with a 5.9\% classification error. The correction head achieves a position MAE of 0.20~mm, roll and pitch MAEs of \(0.27^\circ\) and \(0.29^\circ\), respectively, and a mean \(R^2\) of 0.76. These results indicate that tactile observations contain sufficient information for the learned policy to identify unsuitable contact and estimate corrective motion on unseen plates.

\subsubsection{Online Evaluation Protocol}

We compare four acquisition strategies to isolate the contributions of contact-quality assessment and pose correction: \textbf{Single-press} performs one contact without adaptation; \textbf{Retry-only} uses \(\hat q\) to trigger up to three attempts without changing the contact pose; \textbf{Position-only} additionally applies the predicted position correction while keeping orientation fixed; and \textbf{Full (ours)} jointly corrects position and orientation after rejected contact.

Each strategy performs a press-based scan over all four lines of the ten online-evaluation plates. Depending on line length, Chinese plates require 10--13 trajectory locations per line and English plates require 9--12, yielding 40 scans and 448 trajectory locations per strategy (244 Chinese and 204 English). At each location, an observation with \(\hat q\geq0.80\) is retained; otherwise, the corresponding strategy is applied for up to three attempts. If no attempt is accepted, the highest-scoring observation is retained. The resulting tactile-image--pose pairs are then reconstructed and decoded using the same fixed downstream pipeline for all strategies.

We evaluate local \emph{tactile quality} after contact adaptation and global \emph{tactile reconstruction} after pose-aware fusion. Tactile quality is the percentage of retained tactile images without missing, incomplete, merged, deformed, or spurious dots. Tactile reconstruction is the percentage of Braille cells whose six dot positions all match the physical plate. Finally, \emph{Braille-line accuracy} measures the percentage of decoded lines that exactly match the ground-truth transcription. Tactile quality and tactile reconstruction are computed independently of the language decoder.


\begin{table}[t]
\caption{Online Braille reading results. C/E/All denote Chinese, English, and pooled tactile reconstruction, respectively.}
\label{tab:online_results}
\centering
\scriptsize
\setlength{\tabcolsep}{3.5pt}
\renewcommand{\arraystretch}{1.05}
\begin{tabular}{@{}lcccc@{}}
\toprule
\textbf{Strategy} & \textbf{Att./loc.} & \textbf{Tactile} & \textbf{Tactile} & \textbf{Line} \\
& & \textbf{quality} & \textbf{reconstruction} & \textbf{acc.} \\
& & \textbf{(\%)} & \textbf{C/E/All (\%)} & \\
\midrule
Single-press  & 1.00 & 61.4 & 62.5/65.9/64.0 & 32.5 \\
Retry-only    & 2.36 & 62.9 & 62.5/66.8/64.4 & 32.5 \\
Position-only & 1.84 & 81.9 & 79.6/80.5/80.0 & 60.0 \\
Full (ours)   & \textbf{1.42} & \textbf{94.0} & \textbf{88.9/88.2/88.6} & \textbf{80.0} \\
\bottomrule
\end{tabular}
\end{table}


\subsection{Online Braille Reading Results}
\label{subsec:online_pressing}

Figure~\ref{fig:adaptive_acquisition} illustrates a representative scan using Full. At \(k=1\) and \(k=5\), initially unsuitable tactile observations are rejected and subsequently corrected until readable contact is established.

Table~\ref{tab:online_results} summarizes the online results. Retry-only provides little improvement over Single-press despite requiring substantially more contacts (2.36 versus 1.00 attempts/location), indicating that repeated acquisition without physical correction rarely resolves unsuitable contact. Position-only increases tactile quality from 61.4\% to 81.9\%, while Full further improves it to 94.0\% with only 1.42 attempts/location. The 12.1-point improvement over Position-only demonstrates the importance of correcting contact orientation in addition to position.

The benefit propagates to downstream reconstruction and reading. Full achieves 88.6\% pooled tactile reconstruction and 80.0\% exact line accuracy, compared with 80.0\% and 60.0\% for Position-only, respectively. Since all strategies use the same reconstruction and decoding pipeline, these improvements are attributable to the quality of the tactile observations supplied by contact adaptation.

\subsection{Discussion}\label{subsec:exp_discussion}
The experiments support a contact-first view of robotic tactile Braille reading. Retry-only shows that detecting unsuitable contact is insufficient when the underlying physical interaction remains unchanged, while the gap between Position-only and Full indicates that both translational and rotational contact errors contribute to corrupted Braille-dot geometry. Importantly, Full achieves higher tactile quality with fewer attempts than the correction baselines, suggesting that the improvement results from more effective corrective actions rather than a larger interaction budget. The resulting gains in tactile reconstruction and line accuracy further show that contact adaptation protects the tactile evidence subsequently used for reconstruction.

The current evaluation is limited to rigid, approximately planar PLA Braille plates and a fixed tactile-sensor mounting. The policy is learned from expert-guided demonstrations and allows at most three contact attempts per trajectory location. Future work will investigate broader surface materials and geometries, coupled contact disturbances, different sensor configurations, and comparisons with general tactile-servoing controllers.

\section{Conclusion}
We presented an image-conditioned adaptive-contact framework for robotic tactile Braille reading. From grouped expert demonstrations, the robot learns to assess contact quality and correct the contact pose, enabling unsuitable contact to be actively re-established before tactile reconstruction. Online Braille-reading experiments demonstrate improved contact reliability and downstream Braille reconstruction. These results highlight that reliable tactile reading depends not only on recognizing sensory observations, but also on actively establishing contact that makes the tactile information readable. Future work will investigate more diverse surfaces, coupled contact disturbances, and less structured exploration settings.



\bibliographystyle{IEEEtran}
\bibliography{references}
\end{document}